\documentclass[11pt,a4paper]{article}
\usepackage[hyperref]{eacl2021}
\usepackage{times}
\usepackage{latexsym}
\usepackage{graphicx}
\graphicspath{ {./images/} }

\usepackage{enumitem}
\usepackage{makecell}
\usepackage{comment}
\usepackage[table]{colortbl}
\usepackage{xspace}

\newcommand*{\yoruba}{Yor\`ub\'a\xspace}

\usepackage{microtype}

\aclfinalcopy 

\title{AI4D - African Language Program}

\author{
\begin{minipage}[t]{\textwidth}
\centering
\normalsize
Kathleen Siminyu$^{1}$,
Godson Kalipe$^{2}$,
Davor Orlic$^{3}$,
Jade Abbott$^{4}$,
Vukosi Marivate$^{5}$,\\
Sackey Freshia$^{6}$,
Prateek Sibal$^{7}$,
Bhanu Neupane$^{7}$,
David I. Adelani$^{8}$,
Amelia Taylor$^{9}$,\\
Jamiil Toure ALI$^{2}$,
Kevin Degila$^{2}$,
Momboladji Balogoun$^{2}$,
Thierno Ibrahima DIOP$^{10}$,\\
Davis David$^{11}$,
Chayma Fourati$^{12}$,
Hatem Haddad$^{12}$,
Malek Naski$^{12}$\\

{\footnotesize \normalfont 
$^{1}$AI4D Africa,
$^{2}$Takwimu Lab,
$^{3}$Knowledge for All Foundation,
$^{4}$Retro Rabbit,\\
$^{5}$University of Pretoria,
$^{6}$Jomo Kenyatta University of Agriculture and Technology,\\
$^{7}$UNESCO,
$^{8}$Saarland University,
$^{9}$The University of Malawi,\\
$^{10}$Baamtu Datamation,
$^{11}$TAVODET Youth Development,
$^{12}$iCompass,
\\
} 
\end{minipage}
}

\date{}

\begin{document}
\maketitle
\begin{abstract}
  Advances in speech and language technologies enable tools such as voice-search, text-to-speech, speech recognition  and machine translation. These are however only available for high resource languages like English, French or Chinese. Without foundational digital resources for African languages, which are considered low-resource in the digital context, these advanced tools remain out of reach. This work details the AI4D - African Language Program, a 3-part project that 1) incentivised the crowd-sourcing, collection and curation of language datasets through an online quantitative and qualitative challenge, 2) supported research fellows for a period of 3-4 months to create datasets annotated for NLP tasks, and 3) hosted competitive Machine Learning challenges on the basis of these datasets. Key outcomes of the work so far include 1) the creation of 9+ open source, African language datasets annotated for a variety of ML tasks, and 2) the creation of baseline models for these datasets through hosting of competitive ML challenges.
\end{abstract}

\section{Motivation}

Languages, with their complex implications for identity, cultural diversity, spirituality, communication, social integration, education and development, are of crucial importance for people, prosperity and the planet. People not only embed in languages their history, traditions, memory, traditional knowledge, unique modes of thinking, meaning and expression, but more importantly they also construct their future through them. 

In this context, Language Technologies (LT), greatly contribute to the promotion of linguistic diversity and multilingualism. These technologies are moving outside research laboratories into numerous applications in many different areas. UNESCO’s International Conference Language Technologies for All (LT4All): Enabling Linguistic Diversity and Multilingualism Worldwide\footnote{\href{https://en.unesco.org/LT4All}{International Conference Language Technologies for All (LT4All)}}, organized in December 2019, underlined  spelling/grammar checkers up to speech and speaker recognition, machine translation for text and audio, speech synthesis, and spoken dialogue among others as important areas for enabling linguistic diversity and multilingualism. 
In addition, the Los Pinos Declaration on the Decade of Indigenous Languages (2022-2032)\footnote{\href{https://en.unesco.org/news/upcoming-decade-indigenous-languages-2022-2032-focus-indigenous-language-users-human-rights}{Upcoming Decade of Indigenous Languages (2022 – 2032) to focus on Indigenous language users’ human rights}} calls for the design and access to sustainable, accessible, workable and affordable language technologies and places indigenous peoples at the centre of its recommendations under the slogan \emph{“Nothing for us without us.”}

\subsection{The increasing technological gap}

Apart from applications in automatic translation which have revolutionized communication between people around the world, NLP has been driving significant changes that African languages speakers are often left out of, because their languages are not yet taken into account in mainstream NLP research. Some of those include the use of chatbots for better customer service and improved user experience and other tools like sentiment analysis solutions that help companies better understand their market and efficiently process huge amounts of feedback about their services.

The lack of African languages' datasets is discouraging many NLP practitioners to start from scratch and the task will have to be taken up by African researchers because of the low economic interest that our languages represent for top companies driving changes in NLP. In fact, Quartz Africa explains that those giants are directly concerned with Return On Investment while investing in new languages. It has appeared that the top 100 languages in NLP cover 96 percent of the GDP of the world\footnote{Language GDP coverage was calculated multiplying the number of speakers of a language by the GDP per capita} although they equated to less than 60 percent of the population.\footnote{\href{https://qz.com/africa/1475763/african-languages-are-lagging-behind-when-it-comes-to-voice-recognition-innovations/}{African languages are being left behind when it comes to voice recognition innovation}} Without action, the digital gap between Africa and the rest of the world will keep increasing as far as NLP solutions which are driving changes in healthcare, finance or education, are concerned.

\subsection{NLP solutions in African languages as a motor of socio economic change}
\begin{figure*}[h]
\includegraphics[width=\textwidth]{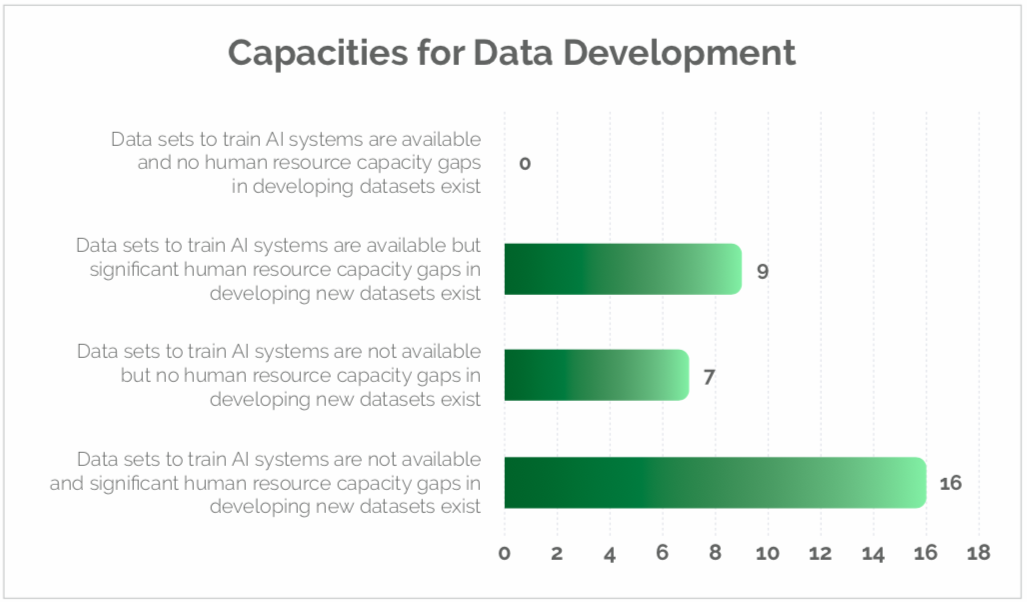}
\caption{Responses of African member states as to their capacities for data development. These were collected as part of an AI Needs Assessment Survey facilitated by UNESCO and AI4D\footnote{\href{https://unesdoc.unesco.org/ark:/48223/pf0000375322/PDF/375322eng.pdf.multi}{Artificial Intelligence Needs Assessment Survey In Africa}}}
\end{figure*}

UNESCO’s report ‘Steering AI and Advanced ICTs for Knowledge Societies’ notes that access to data is an important determinant for the development of AI. For instance, access to data is essential for training algorithms and for their usefulness in large scale application. It notes that “in the absence of access to data, new firms face potential entry barriers in challenging the entrenched market actors”. Further, even academic institutions face barriers in accessing data, which becomes a barrier to research and innovation ~\cite{hu2019steering}. 

In response to UNESCO and AI4D's Artificial Intelligence Needs Assessment Survey in Africa (Figure 1), nine out of thirty  countries in Africa have underlined  the availability of datasets to train AI systems but a lack of human resources for developing datasets.\footnote{Botswana, Cabo Verde, Cameroon, Côte d'Ivoire, Equatorial Guinea, Madagascar, Senegal, Seychelles, Togo} Another 16 do not have datasets to train AI systems nor the capacities to develop new datasets.\footnote{Benin, Chad, Comoros, Democratic Republic of the Congo, Eswatini, Gambia, Guinea, Lesotho, Malawi, Nigeria, Sao Tome and Principe, Sierra Leone, Somalia, Uganda, Zambia, Zimbabwe}

Initiatives have been spurring around the continent to experiment the use of indigenous languages as a means of education in the past years. Studies led by the Unesco Institute for Lifelong Learning have shown that multilingual education in African countries actually yield better results than its monolingual alternative and is more benefitial to socio-economic development ~\cite{ouane2010and}. But, new technological tools are needed to support initiatives such as this. In a wider sense, many of the initiatives for revitalizing African languages have been suffering from lack of funding, documentation and innovative approaches even when national policies are in favor of their use ~\cite{sands2018language}. Technological solutions based on NLP have the ability to reduce program costs for language revitalization programs by helping them reach more ambitious goals, faster and with less human resources.

In general, NLP in Africa holds a huge potential, as advances in NLP for African languages will constitute a win not only for the speakers of those languages but also for companies offering services on the continent. Governments are another potential great beneficiary of breakthroughs in the domain as they would be able to leverage them to bring information and modern nation-wide solutions closer to the inhabitants of the continent who have little or no proficiency in foreign languages.

\section{Literature Review}

Few endeavors in the field of Artificial Intelligence have been working towards facilitating data collection and facilitating NLP research efforts on the African continent.

Masakhane~\cite{orife2020masakhane}, is one of the most prominent recent efforts towards machine translation research in Africa and offers greater visibility of the community on global platforms. They publicly offer easy to grasp hands-on tools to on-board newcomers to the field and allow them to train baseline models for a wide range of African languages using the JW300 dataset ~\cite{agic2020jw300}. They have helped build baseline models for more than 35 African languages to date.
Despite the tremendous work that Masakhane has been doing, there is still a need for more diverse data as JW300 data is heavily biased with religious content. It is one of the motivations of the African Language program. Another concern is the lack of written documentation for most widely spoken African languages. Initiatives such as the BULB Project ~\cite{adda2016breaking} aim at facilitating direct audio data collection for African languages and their annotation at phoneme and word level.

Much earlier efforts included the Language Resources and Evaluations Journal Special Issue on African Language Technology~\cite{de2011introduction} which presented a cross-section of the state-of-the-art in the field as well as an overview of the most prominent research efforts in African Language Technology. In 2016, Interspeech had a special session titled "Sub-Saharan African languages: from speech fundamentals to applications" in recognition of the fact that Sub-Saharan African languages tend to remain under-resourced, under-documented and often also un-written.\footnote{\href{https://www.aflat.org/content/cfp-interspeech-2016-special-session-sub-saharan-african-languages-speech-fundamentals-appli}{Interspeech 2016 Special Session}}

\section{Program Description}
\label{gen_inst}

AI4D - Artificial Intelligence for Development, is an initiative to improve the quality of life for all in Africa and beyond by partnering with Africa’s science and policy communities to leverage AI through high-quality research, responsible innovation, and strengthening talent.
The AI4D - African Language Program was conceptualized as part of a roadmap to work towards better integration of African languages on digital platforms, in aid of lowering the barrier of entry for African participation in the digital economy. 
It was organised in 3 key phases;

\subsection{Language Dataset Challenges}
The AI4D Language Dataset Challenges\footnote{\href{https://zindi.africa/competitions/ai4d-african-language-dataset-challenge}{AI4D Africa Language Challenge}} were framed to focus on data collection, in response to the challenges of low availability of input data for African languages and the poor discoverability of resources that do exist, thus hindering the ability of researchers to do machine translation~\cite{martinus2020neural}, and other NLP tasks. We put together a panel of judges and performed both qualitative and quantitative evaluations, based on datasheets~\cite{gebru2018datasheets} submitted with each datasets and the datasets themselves, to identify outstanding submissions. The two rounds of this challenge yielded 52 dataset submissions accross 15 African languages with 13 winning prizes.

Some of the key observations identified from this phase of the project include \cite{siminyu2020ai4d}:

\begin{itemize}[noitemsep,nolistsep,leftmargin=*]
\item Teams composed of individuals from relevant multi-disciplinary backgrounds, including computer scientists, professional translators and linguists, were able to create and annotate datasets that captured fundamental lexical and semantic nuances of languages.

\item The challenge framing allowed for anyone to participate. While useful as an exercise in evaluating the interest in such a challenge, high quality submissions came from teams who had been exposed to NLP research work. 
\item Since the challenge was evaluated monthly, we often received disparate submissions from the same teams. Instead, one large dataset built over a couple of months would have been the ideal outcome.

\end{itemize}

These observations motivated the design of a subsequent phase of the project, the Fellowship.

\subsection{Language Dataset Fellowships}
From the top teams that participated in the challenges, we invited nine to take part in a subsequent phase of the program, a 3-4 month Fellowship Program\footnote{\href{https://www.k4all.org/project/language-dataset-fellowship/}{Cracking the Language Barrier for a Multilingual Africa}}. This provided research grants for teams to invest in resources, gave enough time for collaborative consultations to determine the sizes of expected datasets, the downstream NLP tasks they would be annotated for as well as mentorship and advisory requirements.

The datasets developed through this process cover a variety of languages and NLP tasks; Machine Translation datasets (Ewe, Fonge, \yoruba, Luganda, Twi and the 11 official languages of South Africa), a Text-to-Speech dataset (Wolof), a Sentiment Analysis dataset (Tunisian Arabizi), a Keyword Spotting dataset (Luganda) and Document Classification datasets (Chichewa and Kiswahili).

The Fellowship also presented a platform to tackle some opportunities identified to support future work in African, and low resource, language dataset creation, including research and analysis of the legal implications of obtaining textual, visual and audio data from a variety of online sources, and the development of copyright, intellectual property and data protection guidelines for NLP researchers.

These guidelines will be published in addition to research papers from the individual fellows on their particular dataset development work.

\begin{table*}[h]
\begin{center}
\begin{tabular}{|l|r|r|r|l|}
\hline \bf Challenge & \bf Participants & \bf Leader Board & \bf Submissions & \bf Best Score \\ \hline
\hline \cellcolor{gray!25}\makecell{Agricultural Keyword Spotter\\ for Luganda} & 713 & 255 & 3385 & 0.650 (Log Loss)  \\ \hline
\hline \cellcolor{gray!25}\makecell{Social Media Sentiment Analysis\\ for Tunisian Arabizi} & 733 & 312 & 6674 & 0.95 (Accuracy)  \\ \hline
\hline Chichewa News Classification & 567 & 208 & 2869 & 0.684 (Accuracy)  \\ \hline
\hline \makecell{French to Fongbe and Ewe\\ Machine Translation} & 240 & 44 & 406 & 0.363 (BLEU)  \\ \hline
\hline \yoruba Machine Translation & 426 & 59 & 443 & 0.468 (BLEU)  \\ \hline
\hline \makecell{Automatic Speech Recognition \\ in Wolof} & 204 & 19 & 76 & 0.110 (WER)  \\ \hline
\hline
\end{tabular}
\end{center}
\caption{\label{font-table} List of challenges and participation metrics where some data is already available. The grey cells indicate where the metrics listed are final as the challenge has been completed.}
\end{table*}

\subsubsection{Data sources} 

The teams used different methods to collect data. Some of the main ones were :

\paragraph{Data scraping from online sources} \hfill \break
Although, it is a general trend that African languages are getting less and less written, there still exist online media content in local languages. Some international media powerhouses such as BBC and Voice Of America have versions of their sites targeted at an African audience with the content exclusively in African languages. In addition, social media remains a place where data collection can be leveraged provided proper measures (automatic or manual) for checking correctness can be put into place. 
Other interesting online sources that were leveraged include TED talks/movies transcripts, radio transcripts and software localization texts~\cite{adelani2021menyo20k}. 

\paragraph{Translators} \hfill \break
The most common way of generating data during the first phase of the program has been using the help of local translators who have had extensive experience translating documents encompassing a wide range of themes and topics. This approach was mostly followed by teams creating datasets for neural machine translation. Data was first generated (sentences created from scratch based on given themes) or collected from online media outlets and blogs \footnote{Consent to used the concerned articles was always sought and obtained before proceeding to data collection} and then shared with translators for translation.

\paragraph{Audio recordings} \hfill \break
Teams building text-to-speech datasets recruited actors to record sentences from the cleaned text datasets using Common Voice \footnote{\href{https://commonvoice.mozilla.org/}{Common Voice}}.

\subsubsection{Challenges in data collection}

There are challenges that are specific to building dataset for low-resourced languages and that have to be expected while attempting to build benchmark datasets. Some of those the fellows encountered had to do with :

\paragraph{The lack of standard in written scripts of the languages} \hfill \break
For a lot of African languages, there are no standardized grammar or orthographic rules. Worse, with these languages being written less and less~\cite{sands2017challenge}, new orthographies have been given to words depending on the country or on the need of the moment. The challenge here is that many resources are encountered each having different orthographies for the same words and different translations for the same types of sentences. To tackle this, researchers reached out to and worked in collaboration with national universities' languages departments and national organizations such as the Ewe Academy in Togo in order make the corpora uniform.

\paragraph{The scarcity of existing resources} \hfill \break
It is the essence of these languages being classified as low-resource languages. Although there is a lot of work being done notably by translators, it is not saved and there is no history of the work that can be accessed which makes years of work volatile.

\paragraph{Intellectual property issues} \hfill \break
Beyond the legal and commercial aspects of intellectual property as it can exist related to any type of work, what can make it particularly hard in an African context is the fact that most authors or organizations that publish content in African languages have little if any online presence allowing one to reach them and request rights to use their work. 
In instances where adequate contact information was available, it was difficult to obtain the permission necessary. We therefore needed to either give up on using these data sources at worst or incurring delays in the work due to formalities necessary to get access to the data.

\paragraph{The scarcity of expert language professionals} \hfill \break
As most African languages are less and less written ~\cite{sands2017challenge}, there is a general loss of academic interest from language students. It makes it particularly hard to encounter linguists, translators, language teachers or other language professionals that are up to date with the latest development of the language. Many attempts might be necessary in order to find suitable language professionals to work with. We have also noticed it is good practice to work with a couple of different professionals in order to cross-validate their work before accepting it.

\paragraph{The reluctance of data holders to give it up} \hfill \break
Data is the new gold and even for those who might not have previously understood this, the sudden interest shown in collecting data from them can raise suspiscions as the concepts of open research are not always clear to some of those data holders who are not familiar to the academic world.

\paragraph{Lack of commitment from partners} \hfill \break
 A major challenge concerning building Text-To-Speech (TTS) datasets consisted in the difficulty to find committed actors to record the text datasets items. The nature of TTS requires the same actor to do all the recordings. Therefore at times, the recordings had to be started over because of the unavailability of the initial actors who started. The feedback has indicated that this had a lot to do with our inability to provide satisfying remuneration rates for the amount of work that was required to generate datasets large enough.

\subsubsection{Data quality}

As far as data quality is concerned, going the extra mile to ensure diacritics were correctly written was a major challenge as it required specialised expertise, was time consuming and therefore demanded that more funds were put into remunerating those working with us or development of automatic diacritics application models ~\cite{orife2018adr}~\cite{orife2020improving}. 
In addition, most African languages are now commonly spoken with the intervention of many borrowed words from the respective countries' official languages (French, English). Cleaning was therefore necessary to keep the text as pure as possible. This was done using language detectors and regular expressions before finalizing with manual cleaning.

Another major quality check concerned the balance of the data. Some datasets contained a huge amount of data of one type or source making it imbalanced and likely to introduce bias in the models built using it. This is a consideration that is not specific to African language datasets and that different fellows have tried to remain aware of while choosing their data sources in order to maintain a proper balance in each dataset.

Also, in order to ensure proper classification categories, language professionals were involved even in data annotation for classification datasets. This was in instances where only translations of titles of news articles were available but did not provide enough information to accurately associate a category with the text.

\subsection{Machine Learning Competitions}
This final phase involved design and split of the datasets into train, development and test sets; the preparation of datasheets to document the motivation, composition, collection process and recommended uses of the datasets and hosting ML competitions on Zindi, an African data science competition platform, so as to engage the wider NLP community.

Six competitions and one hackathon were hosted on Zindi as follows:
\begin{itemize}[noitemsep,nolistsep,leftmargin=*]
\item AI4D iCompass Social Media Sentiment Analysis for Tunisian Arabizi\footnote{\href{https://zindi.africa/competitions/ai4d-icompass-social-media-sentiment-analysis-for-tunisian-arabizi}{Social Media Sentiment Analysis for Tunisian Arabizi}} 
\item AI4D Yorùbá Machine Translation Challenge\footnote{\href{https://zindi.africa/competitions/ai4d-yoruba-machine-translation-challenge}{Yorùbá Machine Translation Challenge}}
\item AI4D Takwimu Lab - Machine Translation Challenge: French to Fongbe and Ewe\footnote{\href{https://zindi.africa/competitions/ai4d-takwimu-lab-machine-translation-challenge}{French to Fongbe and Ewe Machine Translation}}
\item AI4D Chichewa News Classification Challenge\footnote{\href{https://zindi.africa/competitions/ai4d-malawi-news-classification-challenge}{Chichewa News Classification}}
\item AI4D Baamtu Datamation - Automatic Speech Recognition in WOLOF\footnote{\href{https://zindi.africa/competitions/ai4d-baamtu-datamation-automatic-speech-recognition-in-wolof}{Automatic Speech Recognition in WOLOF}}
\item AI4D Swahili News Classification Challenge \footnote{\href{https://zindi.africa/hackathons/ai4d-swahili-news-classification-challenge}{Swahili News Classification Challenge}}
\item GIZ NLP Agricultural Keyword Spotter for Luganda\footnote{\href{https://zindi.africa/competitions/giz-nlp-agricultural-keyword-spotter}{Agricultural Keyword Spotter for Luganda}}
\end{itemize}

More challenges are to come as the data preparation gets completed. The objective of these competitions is to :

- Generate benchmark models that can be built upon by other researchers and that can assist local languages professionals in their tasks.

- help create tools that can be used in insurance companies, banking or social media mining to better understand users' feelings.

- Assist material creation for education in African languages, bring services such as banking or healthcare closer to local languages speakers among many other possible use cases.


\section{Future Scope of the project}

\subsection{Achievements of the program}
The program has been successful in creating a grass roots movement of interdisciplinary researchers and professionals across Africa to collaborate on addressing the challenge of lack of access to data in African languages that may inhibit the participation of people speaking these languages in the digital
economy.
The program has been highlighted at the launch of the Open for Good Alliance, an initiative to support the development of localized training data for AI-driven innovation\footnote{\href{https://en.unesco.org/news/unesco-organises-workshop-strengthening-multilingualism-through-datasets-low-resourced}{UNESCO organises workshop on strengthening multilingualism through datasets in low resourced languages at IGF 2020}}. The Alliance was launched in November 2020 by the International Development Research Centre, FAIR Forward / GIZ, Mozilla Foundation, Radiant Earth Foundation, \textless A+\textgreater Alliance for Inclusive Algorithms, African Institute for Mathematical Sciences, Kwame Nkrumah University of Science
and Technology, Makerere University and UNESCO with three objectives:
\begin{itemize}[noitemsep,nolistsep,leftmargin=*]
\item Making training datasets openly available, helping existing open training datasets to be found and supporting their maintenance.
\item Facilitating the coordination and exchange of good practices and ideas through discussions with community members around the collection of datasets and applications of AI, supporting the development of
standards where needed, and share successful examples.
\item Increasing the public awareness for the benefits of openly available, unbiased and localized training data.
\end{itemize}
This program will also be used as a model case to inform evidence-based policy making concerning Artificial Intelligence and will be included in UNESCO’s AI Decision maker’s Essential to inform policy makers. The generated datasets have been hosted on Zenodo in a community created for African NLP\footnote{\href{https://zenodo.org/communities/africanlp}{African Natural Language Processing (AfricaNLP)}} where we hope to encourage other enthusiasts to use them for Machine Translation, Text Classification, Named Entity Recognition, overloaded terms or loan words from English detection and also comparative vocabulary analysis.

This fellowship has been an opportunity for the fellows to build networks of language professionals and institutions that collaboration will go on with in order to produce better datasets in terms of size, quality, annotations, cleanliness and diversity among other criteria. More funding would be required to take the work further as professional expertise in African languages is becoming a rare skill. In addition, due to the time consuming nature of tasks such as translation, which can get even more complex when dealing with languages with diacritics, fuzzy grammar and orthographic standards, working with African languages is expensive.

\subsection{Future challenges}

One major challenge moving forward will be the creation of datasets and solutions that capture the intrinsic multilingual nature of modern African communication and the fact that code-switching is very common in daily communication. Efficient NLP for African languages could look like; mixed language processing tools that can catch up with the democratization of English and Swahili coming together in Kenya’s Sheng, the mixing of French with Arabic or Berber in Algeria, or even the Romanization of Amharic.\footnote{\href{https://qz.com/africa/1475763/african-languages-are-lagging-behind-when-it-comes-to-voice-recognition-innovations/}{African languages are being left behind when it comes to voice recognition innovation}}

Another challenge remains in the form of ways in which the momentum generated by the project(s) can be leveraged in scaling up the work, networks and impact of the program. Possible pathways to achieve scale require organising the network around frameworks that can guide the community in developing and sustaining data commons.

\subsubsection{Data Commons Frameworks}

In this respect, a Data Commons Framework would be one relevant model to organize future work around. There have been various frameworks proposed, such as this Data Commons Version 1.0 Framework\footnote{\href{https://medium.com/berkman-klein-center/data-com- mons-version-1-0-
a-framework-to-build-toward-ai-for-good-73414d7e72be}{Data Commons Version 1.0: A Framework to Build Toward AI for Good}} proposed by the Berkman Klein Centre which is intended to build towards AI for Good.

Expanding access to data is crucial for the development of machine learning and AI. Yet even as data commons develop, other concerns related to representativeness of the data, discrimination and openness need to be addressed. 

\subsubsection{Framework for participatory development of language datasets}

In order to successfully scale up, the programme must expand its reach among different actors engaged in language development to tap into interdisciplinary knowledge and resources. As a tool for participatory language technology development, the Masakhane research community in ~\cite{nekoto2020participatory} details a framework it has developed with entry points for content creators, translators, curators, language technologists and evaluators to work together in addressing some of the challenges of language development.
Such a framework would enable the program to focus on efforts in expanding the scope of its research in terms of quality, quantity and impact as well as involve other stakeholders in the language ecosystem to take part in the development stages of AI language tools. 

\subsubsection{Near term needs of the programme}
Near term requirements over a period of the next 1 to 2 years include:
\begin{itemize}[noitemsep,nolistsep,leftmargin=*]
    \item Funding support to sustain the fellowships and expand the work on datasets so as to enable the development of minimum viable products based on these datasets
    \item Funding support to develop technical infrastructure for hosting the data commons, including to ensure that the data is \textbf{F}indable, \textbf{A}ccessible, \textbf{I}nteroperable and \textbf{R}eusable in line with the FAIR principles\footnote{\href{https://www.go-fair.org/fair-principles/}{FAIR Principles}}
    \item Research on legal and policy issues concerning data protection, data sharing and transfer within different jurisdictions to strengthen the collaborative work through the program 
    \item Enlarging the network of stakeholders involved in the project to facilitate greater multi-stakeholder cooperation 
    \item Mentoring and engaging young researchers through continent wide hackathons, workshops, trainings and conferences
    \item Advocacy towards governments and policymakers to garner support for the programme at the national, regional and global level.
 \end{itemize}
 
\section*{Acknowledgements}
This work has been sponsored through a partnership between several organisations, listed below in alphabetical order;
\begin{itemize}[noitemsep,nolistsep,leftmargin=*]
\item The AI4D Africa Initiative
\item The Centre for Intellectual Property and Information Technology(CIPIT), Strathmore University
\item The Data Science for Social Impact Research Group, University of Pretoria
\item Deutsche Gesellschaft für Internationale Zusammenarbeit(GIZ)
\item The International Development Research Centre(IDRC)
\item The Knowledge 4 All Foundation(K4All)
\item Masakhane
\item The Swedish International Development Cooperation Agency(Sida)
\item The United Nations Educational, Scientific and Cultural Organization(UNESCO)
\item Zindi Africa
\end{itemize}

\bibliography{anthology,eacl2021}
\bibliographystyle{acl_natbib}

\end{document}